\documentclass{article}
\usepackage{ijcai21}
\usepackage{times,soul,url,amsmath,amsthm,algorithm,algorithmic,amsfonts,bm,booktabs,tikz,pgf,pgfplots,amsmath,soul,url,graphicx,amsmath,booktabs,booktabs,nicefrac,bm,colortbl,tikz,pgfplots,enumerate,xargs,optidef}
\usepackage[hidelinks]{hyperref}
\usepackage[utf8]{inputenc}
\usepackage[small]{caption}
\usetikzlibrary{arrows,automata,circuits,arrows,plotmarks,decorations.markings,trees,shapes}
\tikzstyle{nodo}=[ellipse,draw=black!100,fill=black!30,line width=.7pt,minimum width=1.2cm,minimum height=.7cm]
\tikzstyle{Qnodo}=[ellipse,draw=black!100,fill=black!10,line width=.7pt,minimum width=1.2cm,minimum height=.7cm]
\tikzstyle{arco}=[draw=black!80,line width=.7pt, postaction={decorate}, decoration={markings,mark=at position 1.0 with {\arrow[ draw=black!80,line width=.7pt]{>}}}]
\tikzstyle{decision} = [rectangle, draw, fill=black!100,text=white, text width=4.5em, text badly centered, node distance=3cm, minimum height=3em]
\tikzstyle{block} = [rectangle, draw, fill=blue!20, text width=5em, text centered, rounded corners, minimum height=3em]
\tikzstyle{line} = [draw, -latex']
\tikzstyle{cloud} = [draw, ellipse,fill=red!20, node distance=3cm, minimum height=2em]

\pgfplotsset{legend image with text/.style={
legend image code/.code={%
\node[anchor=center] at (0.3cm,0cm) {#1};}},}
\pgfplotsset{compat=1.13}

\title{ADAPQUEST: A Software for Web-Based Adaptive Questionnaires\\based on Bayesian Networks}
\author{Claudio Bonesana \and Francesca Mangili \and Alessandro Antonucci
\affiliations
Istituto Dalle Molle di Studi sull'Intelligenza Artificiale (IDSIA) - Lugano, Switzerland
    \emails
    {\tt\{claudio.bonesana,francesca.mangili,alessandro\}idsia.ch}}
\begin{document}
\maketitle
\begin{abstract}
We introduce ADAPQUEST, a software tool written in Java for the development of adaptive questionnaires based on Bayesian networks. Adaptiveness is intended here as the dynamical choice of the question sequence on the basis of an evolving model of the skill level of the test taker. Bayesian networks offer a flexible and highly interpretable framework to describe such testing process, especially when coping with multiple skills. ADAPQUEST embeds dedicated elicitation strategies to simplify the elicitation of the questionnaire parameters. An application of this tool for the diagnosis of mental disorders is also discussed together with some implementation details.
\end{abstract}
\section{Introduction}\label{sec:intro}
A questionnaire is called \emph{adaptive} when its question sequence is dynamically driven by the answers of the taker. The typical goal is to optimally estimate an aspect of interest of the test taker described by a set of target variables (e.g., his/her skills) while also reducing as much as possible the number of questions.

Algorithm~\ref{alg:ta} depicts a standard workflow for adaptive questionnaires. The best question ($Q^*$) to ask in a particular stage of the questionnaire is picked from an item pool ($\bm{Q}$) by a function (${\tt Pick}$) of the previous answers ($\bm{e}$). The answer of the test taker ($\sigma$) is consequently collected (${\tt Answer}$) and the process iterated unless some (${\tt Stopping}$) condition, still based on the previous answers, is achieved. Finally a function (${\tt Evaluate}$) returns a grade based on all the answers collected before the end of the questionnaire.

\begin{algorithm}[htp!]
\begin{algorithmic}[1]
\STATE $\bm{e}\gets\emptyset$
\WHILE{ {\bf not} $\tt{Stopping}(\bm{e})$}
\STATE $Q^* \gets {\tt{Pick}}(\bm{Q},\bm{e})$
\STATE $q^* \gets {\tt{Answer}}(Q^*,\sigma)$
\STATE $\bm{e} \gets \bm{e} \cup \{ Q^*=q^* \}$
\STATE $\bm{Q} \gets \bm{Q} \setminus \{ Q^*\}$
\ENDWHILE
\STATE {\bf return} $\tt{Evaluate}(\bm{e})$
\end{algorithmic}
\caption{Adaptive questionnaire workflow: given student $\sigma$ and item pool $\bm{Q}$, a grade based on answers $\bm{e}$ is returned.\label{alg:ta}}
\end{algorithm}

Trading off accuracy and the number of questions is the typical challenge with real adaptive systems. Algorithms to drive the selection mechanism are extremely important to improve the quality and the reliability of the evaluation process in modern interactions with users. This has to be supported by flexible interfaces able to provide such adaptiveness and interact with portable implementations of the above algorithms.

An important field of application of adaptive questionnaires is education, where they can be used both for training and assessment. In classical assessment tests, tools to achieve some form of adaptiveness by simple deterministic rules have been considered \cite{devellis2006classical}. More successful results can be achieved by the latent modelling of the skill level of the taker \cite{courville2004empirical}.

\emph{Item response theory} (IRT) is the most popular approach of this kind \cite{embretson2013item}. The probability of a correct answer is described by a logistic model with a small number of parameters. Under standard independence assumptions, this allows for a simple updating process, thus making also very easy the implementation of adaptive strategies. As a matter of fact, implementing IRT, but also more  sophisticated techniques, such as the Rasch model \cite{brinkhuis2020dynamic}, in a computer system is relatively straightforward and a huge number of tools for e-learning tools currently embed these algorithms.\footnote{E.g., {\tt\href{http://www.concertoplatform.com}{concertoplatform.com}}.}

Despite its popularity and advantages at the implementation level, IRT might be unable to properly cope with questionnaires aiming to evaluate multiple target variables in the same moment \cite{millan2000using}. The IRT independence assumptions in those cases might be unrealistic, and the model consequently performs poorly. In order to overcome these weaknesses, other formalisms have been considered. Among them, Bayesian networks (BNs, \cite{koller2009}) emerged as a sensible choice able to guarantee an accurate selection of the items \cite{vomlel2004bayesian}, but also a good explainability of the actions \cite{almond2015bayesian}. 

However, in spite of a huge amount of adaptive tools based on IRT, BNs are much less used in this area. To the best of our knowledge, the software we are presenting, called ADAPQUEST\footnote{See {\tt\href{http://www.github.com/IDSIA/adapquest}{github.com/IDSIA/adapquest}}.} is the first mature contribution of this kind. 

We see two possible explanations for such situation. First, although BN inference is nowadays a standard technique, the number of freely available libraries for this task is limited and their embedding in other software projects might be not smooth, while an implementation from scratch would require dedicated efforts not always compatible with an application project. Second, the number of parameters to be tuned for a BN approach might be large and typically higher than those needed for IRT. As the target variables are often regarded as latent variables, learning them from data might not be possible (see \cite{plajner2020monotonicity} for dedicated data approaches) and elicitation techniques should be considered instead. This might be time consuming and also complicated for practitioners not confident with probabilistic graphical models, thus preventing a widespread diffusion of those flexible approaches.

Such situation motivated us to present ADAPQUEST, as a new freely available Java software tool embedding BN inference and modelling features implemented for the design of adaptive tests and their practical implementation through web interfaces. ADAPQUEST supports state-of-the-art techniques for both the elicitation process, intended to make as simple and as smooth as possible such elicitation process, and the adaptive selection of the items intended to guarantee the necessary explainability of the process \cite{antonucci2021}. The tool is directly built on the top of a recently developed library for probabilistic graphical models, that takes care of the necessary inference tasks \cite{huber2020a}.

The paper is organised as follows. In Section \ref{sec:bn} we discuss the basic ideas of adaptive testing based on BNs and the tools used for explainability and elicitation. In Section \ref{sec:sw} we give some technical information related to the development of ADAPQUEST. In Section \ref{sec:app} we consider a case study, already implemented in ADAPQUEST, and  freely available to the community as a demonstrative project. A discussion about possible outlooks concludes the paper in Section \ref{sec:conc}.

\section{Adaptive Questionnaires by Bayesian Nets}\label{sec:bn}
Bayesian networks (BNs) \cite{koller2009} are a popular class of probabilistic graphical models designed for a compact specification of joint, generative, probability distributions. To implement adaptive questionnaires with BNs we regard the set of \emph{skills} to be evaluated during the questionnaire as a joint variable $\bm{S}$, and we similarly regard the item pool $\bm{Q}$ as a set of variables. Here we only consider discrete variables. The BN uses a directed acyclic graph over these variables as a model of the conditional independence relations among them. This allows for a compact specification of the distribution $P(\bm{S},\bm{Q})$. Given such a generative model, standard algorithms for BNs can be used to answer queries about the model variables. E.g., $P(\bm{S}|\bm{e})$ is the posterior probability for the skills given the answers $\bm{e}$ to the questions in $\bm{Q}$ properly asked to the taker. This distribution can be used to grade the taker at the end of the questionnaire, but also to decide whether or not to keep asking questions. For the latter task information-theoretic measures, such as the entropy $H(\bm{S}|\bm{e})$ of the posterior distribution over the skills given the answers received so far, are used and we typically stop the questionnaire when this entropy level goes below a threshold $H^*$. The selection of the optimal question to ask to the taker is slightly more involved: as the actual answer to a candidate question $Q$ is not known, expectation obtained by a weighted average over the probability for the possible answers (i.e., the conditional entropy $H(\bm{S}|Q,\bm{e})$) should be considered instead. To detect the optimal question we maximize the \emph{information gain}, i.e., the difference between the current entropy and the conditional one for the candidate question. Algorithm \ref{alg:bnta} depicts a typical workflow for BNs. The final grade is also an expectation, based on the posterior distribution of a function $f$ able to precisely grade the taker when no uncertainty about the skills is present.

\begin{algorithm}[htp!]
\begin{algorithmic}[1]
\STATE $\bm{e}=\emptyset$
\WHILE{$H(\bm{S}|\bm{e}) > H^*$}
\STATE $Q^* \gets \arg\max_{Q \in \bm{Q}} \left[ H(\bm{S}|\bm{e})-H(\bm{S}|Q,\bm{e}) \right]$
\STATE $q^* \gets {\tt{Answer}}(Q^*,\bm{s}_\sigma)$
\STATE $\bm{e} \gets \bm{e} \cup \{ Q^*=q^* \}$
\STATE $\bm{Q} \gets \bm{Q} \setminus \{ Q^*\}$
\ENDWHILE
\STATE {\bf return} $\mathbb{E}_{P(\bm{S}|\bm{e})}[f(\bm{S})]$
\end{algorithmic}
\caption{Adaptive questionnaire workflow based on a BN over the questions $\bm{Q}$ and the skills $\bm{S}$: given the taker profile $\bm{s}_\sigma$, the algorithm returns an evaluation corresponding to the expectation of an evaluation function $f$ with respect to the posterior for the skills given the answers $\bm{e}$.\label{alg:bnta}}
\end{algorithm}

The procedure is extremely easy to achieve, provided that a reliable BN inference engine to compute $P(\bm{S}|\bm{e})$ and $P(Q|\bm{e})$ is available. In terms of explainability the model offers high transparency: the numerical values leading to a particular question selection, to the stopping condition and to a grade might be reported online during the test execution. Moreover, the techniques recently proposed in \cite{antonucci2021} allow to associate with such quantitative information the modal state of the variables, this providing a qualitative summary of the different actions.

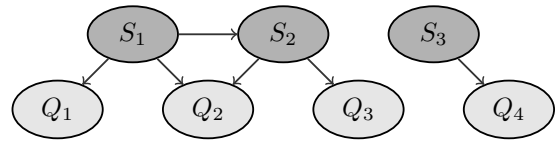
\begin{figure}[htp!]
\centering
\begin{tikzpicture}[scale=1]
\node[nodo] (s1)  at (0,1) {$S_1$};
\node[nodo] (s2)  at (2,1) {$S_2$};
\node[nodo] (s3)  at (4,1) {$S_3$};
\node[Qnodo] (q1)  at (-1,0) {$Q_1$};
\node[Qnodo] (q2)  at (1,0) {$Q_2$};
\node[Qnodo] (q3)  at (3,0) {$Q_3$};
\node[Qnodo] (q4)  at (5,0) {$Q_4$};
\draw[arco] (s1) -- (s2);
\draw[arco] (s1) -- (q1);
\draw[arco] (s1) -- (q2);
\draw[arco] (s2) -- (q2);
\draw[arco] (s2) -- (q3);
\draw[arco] (s3) -- (q4);
\end{tikzpicture}
\caption{BN for questionnaires with three skills and four questions.}
\label{fig:minicat}
\end{figure}

The only critical part of such workflow is the learning of the BN structure and its parameters. In principle, given a data set of observations for $\bm{S}$ and $\bm{Q}$, standard statistical learning techniques could be used. This typically requires complete data, but algorithms to cope with partially incomplete data are also available. Yet, the skills $\bm{S}$ are typically represented by latent variables and their observations unavailable. This particular situation when modelling adaptive questionnaires have been investigated, and specialised learning techniques have been developed (e.g., see \cite{plajner2020monotonicity}).

In ADAPQUEST, we assume the design of the questions and the quantification of the BN to take place in the same time. Thus, when no complete data are available, we assume the quantification process to be based on an elicitation process from a domain expert (e.g., the instructor). Regarding the structure, the natural interpretability of the directed graph underlying a BN makes this task simple: first a graph over $\bm{S}$ is elicited in order to reflect the dependencies between the skills (e.g., in Figure \ref{fig:minicat}, the first two skills are connected and hence dependent, while the third is independent from the first two). Regarding the questions, as we assume them to correspond to the children of the skills, the only effort is to identify the skills relevant to properly answer a particular question.

Once this qualitative part has been achieved, the quantification of the parameters in a BN corresponds to assessing probabilities for single nodes/variables given all the values of the parents. Typical questions are about the probability of a correct answer if the taker has the necessary skills. Or the probability of having skill $S_2$ while not having $S_1$, and so on.

Alternative parametrisations has been proposed to make such elicitation easier \cite{antonucci2021}. Consider for instance a Boolean skill $S$ and a Boolean question $Q$. Two probabilities such as $p:=P(q|s)$ and $p':=P(q|\neg s)$ corresponding to the probability of a correct answer given that the taker has or has not the skill are sufficient to quantify the relation. A possibly simpler parametrisation is provided by $\delta:=p-p'$ and $\gamma:=1-\frac{p+p'}{2}$, the two numbers being still normalised between zero and one and giving the discriminative power of the question and its difficulty. These parameters can be extended to the general case, and make faster and easier the elicitation efforts when adding new items to the pool.

\section{The ADAPQUEST Software}\label{sec:sw}
ADAPQUEST is a REST micro-service written in Java using the Spring Boot framework.\footnote{\href{https://spring.io/projects/spring-boot}{See \tt spring.io/projects/spring-boot}.}
The structure is extremely flexible and the configuration of a new survey/test can be done via API, via code, or by using simple JSON files. If needed, the tool can also connect to an already configured and compatible database where the item pool and the model are stored. The simple hierarchy of classes and a state-of-the-art architecture allows the system to be easily expanded with new features and adaptive criteria to cover further explorations on the field.

From a developer point of view, each questionnaire/survey is composed by three parts: (i) a BN model used to perform inference in order to find the next best question based on the history of given answers as in Algorithm \ref{alg:bnta}; (ii) a pool of questions, associated with their BN nodes; (iii) and an highly configurable part that act on the adaptive engine. The BN inferences are based on the CreMA library \cite{huber2020a}. 

The tool is intended to be a back-end service that needs a custom front-end web-application to show the questions. Yet, a demonstrative web interface is already available. This can be used to test the functioning of the tool itself, of new models, and the questions flow. An exchange library can be used to query and manage a remote tool from a client, allowing standard CRUD operations on the surveys and the stored answers. This also allows to integrate the tool as a library inside other projects and easily interact with it. This is especially important when running extensive simulations.

\section{A Case Study on Mental Disorder Diagnosis}\label{sec:app}
Finally, let us present a case study involving the practical use of ADAPQUEST in the development of a survey. The code for the specification of the model used for that is freely available in the ADAPQUEST repository and can be used as a guidance for the development of new applications within this framework. Although here the focus is on a survey, the implementation of a test would be identical. We refer the reader to \cite{mangili2017b} for an educational application whose items cannot be disclosed for the sake of confidentiality. 

Following the empirical evidence corroborating the relation between job quality and mental health \cite{bracci2020perceived}, ADAPQUEST was applied to the development of a questionnaire for the early detection of employees at risk of mental health problems. Data for the development of the model were taken from the Swiss Household Panel (SHP) \cite{tillmann2016swiss}. The association between job quality and health being the focus of the questionnaire, mostly job related questions were selected from the panel. General demographic information (e.g., age, region, education, etc.) and questions about the current perceived mental states (e.g., self assessed degree of anger, happiness, etc. on a scale from 0 to 10) were also included. Overall, the questions database contains 48 questions based on which the risk of developing a stated of distress, experiencing lack of happiness or running into psychological disorders within one year are estimated. Those three events of interest are described by the binary variables \textit{distress}, \textit{lack}, \textit{disorder}.

The BN model used in the adaptive questionnaire is based on a naive Bayes classifier \cite{koller2009}, the classes being represented by the joint state of the three variables \textit{distress}, \textit{lack} and \textit{disorder}, which is described by a single node called \textit{target}. After collecting a sufficient number of answers, the system computes from the posterior probabilities of the \textit{target} node the marginal probabilities of the three health-related variables being true. Such probabilities are taken as a measure of risk for the mental well-being of the test taker. The naive Bayes model adopts the questionable assumption of independence of the question nodes given the target node, however, the predictive performance of that model exceeded that of more complex networks. Namely, with a 10-fold cross validation over the SHP data set (including 57422 instances) we estimated an AUC of 0.78 (standard deviation = 0.01) 0.88 (s.d. = 0.01) and 0.78 (s.d. = 0.03), for the \textit{distress}, \textit{lack} and \textit{disorder} variables, respectively.

\section{Conclusions and Outlooks}\label{sec:conc}
We presented a new software tool for the design and implementation of adaptive questionnaires and surveys based on Bayesian networks. The tool is freely available to the community. As a necessary future work we intend to support \emph{credal} networks \cite{piatti2010a} for the design of adaptive questionnaires. This can be based on the ideas outlined in \cite{mangili2017b} and later extended by \cite{antonucci2021}. This would allow to support interval-valued elicitation, thus providing higher realism in the modelling step.

\newpage 
\section*{Acknowledgments} {The case study has been realised using the data collected by the SHP, which is based at the Swiss Centre of Expertise in the Social Sciences FORS. The project is financed by the Swiss National Science Foundation}
\bibliographystyle{named}
\bibliography{main}
\end{document}